\documentclass{llncs}
\usepackage{makeidx}
\usepackage{blindtext}
\usepackage{graphicx}
\usepackage{makeidx}
\usepackage{float}
\usepackage{multirow}
\usepackage{subfig}
\usepackage{adjustbox}
\usepackage{makeidx}
\usepackage{float}
\usepackage{multirow}
\usepackage{subfig}
\usepackage{adjustbox}

\begin{document}
\frontmatter          
\pagestyle{headings}  
\pagenumbering{arabic}
\addtocmark{Brain Tumor Segmentation from Multi Modal MR images using Fully Convolution Neural Network} 

\title{Automatic Segmentation and Overall Survival Prediction in Gliomas using Fully Convolutional Neural Network and Texture Analysis}
\titlerunning{Automatic Segmentation and Overall Survival Prediction in Gliomas using Fully Convolutional Neural Network and Texture Analysis}  
%
\author{Varghese Alex  \and Mohammed Safwan \and 
Ganapathy Krishnamurthi}
\authorrunning{Alex et al.} 
%
%
\institute{Indian Institute of Technology Madras, Chennai, India,\\
\email{gankrish@iitm.ac.in}}

\maketitle              


\begin{abstract}
In this paper, we use a fully convolutional neural network (FCNN) for the segmentation of gliomas from Magnetic Resonance Images (MRI). A fully automatic, voxel based classification was achieved by training a 23 layer deep FCNN on 2-D slices extracted from patient volumes. The network was trained on slices extracted from 130 patients and validated on 50 patients. For the task of survival prediction, texture and shape based features were extracted from T1 post contrast volume to train an XGBoost regressor. On BraTS 2017 validation set, the proposed scheme achieved a mean whole tumor, tumor core \& active dice score of 0.83, 0.69 \& 0.69 respectively and an accuracy of 52\% for the overall survival prediction. 
%
\end{abstract}
\section{Introduction}

In this paper, we utilize a 23 layer deep FCNN for the task of segmentation of gliomas from MR scans. In the field of medical image analysis, U-net  \cite{unet} is one of the oft used architecture. The network used in this work has a similar architecture as that of the aforementioned network. The network was trained on 2-D axial slices (240 $\times$ 240) extracted from FLAIR, T2, T1, and T1 post contrast sequences. The architecture of the network enables semantic segmentation i.e. classify of all voxels in a slice in a single forward pass. Due the reason stated above, the inference time associated with FCNN based network are lower when compared to traditional patch based CNNs.

\par Convolutional neural network and its variants being deterministic approaches tend to mis-classify voxels as lesion in regions like brain stem, cerebellum where occurrence of gliomas is anatomically impossible. We utilize 3-D connected component analysis to discard components below a certain threshold for false positive reduction. Unlike previous years BraTS competitions, apart from segmentation of gliomas from MR volumes, an additional challenge of predicting the prognosis of subject using pre-operative scan was made part of the BraTS 2017 challenge. 

\par For the prognosis challenge, the overall survival rate was categorized into three groups namely short survivors (prognosis, $p$ $<$ 10 months), mid survivors (10 months $<$ $p$ $<$ 15 months)  \& long survivors ($p$ $>$ 15 months).  The segmentations produced by the FCNN was used to extract first order texture and shaped based features such as entropy, skewness, circularity of lesion constituent etc. The extracted features along with age of the subject was fed to an Extreme gradient boosting (XGBOOST) regressor to predict the prognosis of the subject.


\section{Materials and Methods}

The proposed technique comprises of following stages:
\begin{enumerate}
    \item Pre-processing of data
	\item Segmentation of gliomas using FCNN
    \item Post-processing using 3-D connected components
    \item Feature extraction for survival rate prediction.
   \item Prediction of survival rate using XGBoost Regressor.
\end{enumerate}

\begin{figure}
\includegraphics[width=\textwidth,keepaspectratio]{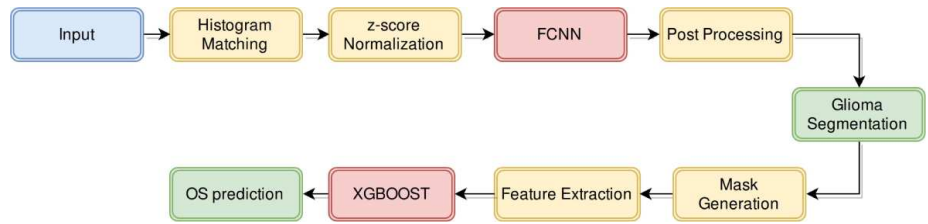}
\caption{Flow chart of the proposed technique}
\label{fig:FLOW}
\end{figure}

The flowchart of the proposed technique is given in Fig. (\ref{fig:FLOW}).
\subsection{Data}

The network was trained and validated on the BraTS 2017 training data \cite{bakas1},\cite{bakas2}, \cite{bakas3},\cite{bakas4}. The training data comprises of 210 HGG volumes and 75 LGG volumes collected from multiple centers. Each patient comprises of FLAIR, T2, T1, T1 post contrast and the associated ground truth labeled by experts. Each sequence was skull stripped and was  re-sampled to 1mm $\times$ 1mm $\times$ 1mm (isotropic resolution).

\par For the overall survival challenge, age \& prognosis of the patient post treatment were supplied by the organizers. The training set for the challenge comprised of 163 High Grade Glioma patients of which 43 patients had survival rate between 10 and 15 months (mid survivors), while 65 patients had prognosis less than 10 months (short survivors) and 56 patients had prognosis greater than 15 months (long survivors).

\subsection{Fully Convolutional Neural Network}
A typical FCNN comprises of convolution operations, max pooling layers and transpose convolution layers. The absence of fully connected layers in FCNNs reduces the number of parameters in the network  \& in-turn accepts inputs of arbitrary sizes. The max pooling layer helps in reducing the spatial dimension of the feature maps in the deeper layers and also aids in capturing translational invariant features in the data. 
\par The dimensionality of the feature maps are brought back to size of the input by either using up-sampling modules such as bilinear interpolation of feature maps or transposed convolution. The use of transposed convolution in the networks makes the scaling procedure of feature maps a parameter to be learned during the training process. Concatenation of feature maps between different layers of the network enables the classifier in the network to make use of both low and complex level features  for better classification results.  
\par FCNNs have an inherent advantage of classifying all pixels in the image by using single forward pass of the image and thus makes FCNNs an ideal choice for semantic segmentation related tasks. Similar to traditional CNNs, the parameters of the network are learned by minimizing the cross entropy.

\section{Preprocessing of Data} 
\subsection{Histogram Matching}
\begin{figure}
\subfloat[]{\includegraphics[width=0.5\textwidth]{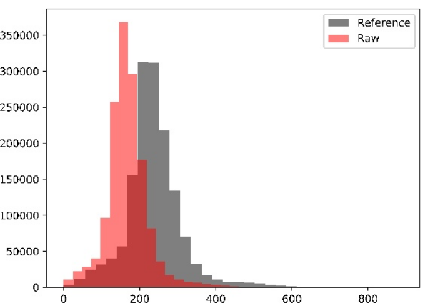}}\hfill
\subfloat[]{\includegraphics[width=0.5\textwidth]{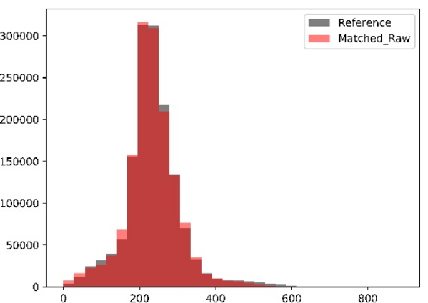}}\\
\caption{Histogram Matching. (a) Raw histograms of reference volume \& test volume. (b) Histograms of reference \& test volume post histogram matching.}
\label{fig:hist_match}
\end{figure}

Multi center data and magnetic field inhomogeneities contribute to the non-uniform intensity variation in MR images. The voxel intensities of all volumes were standardized by matching histograms to an arbitrary chosen reference image from the training database, (Fig.\ref{fig:hist_match}).
\subsection{Z-score Normalization}

The histogram matched volumes were normalized to have zero mean and unit variance using Eq.\ref{eq:norm}, where $X$ is the MR volume, $\mu$ and $\sigma$ are the mean and standard deviation of the volume and $X_{norm}$ is the normalized volume.
\begin{equation}
X_{norm}= \frac{X-\mu}{\sigma}
\label{eq:norm}
\end{equation}

\section{Segmentation of Gliomas using proposed network}

\subsection{Network Architecture}

\begin{figure}
\subfloat[]{\includegraphics[width=0.6\textwidth,keepaspectratio]{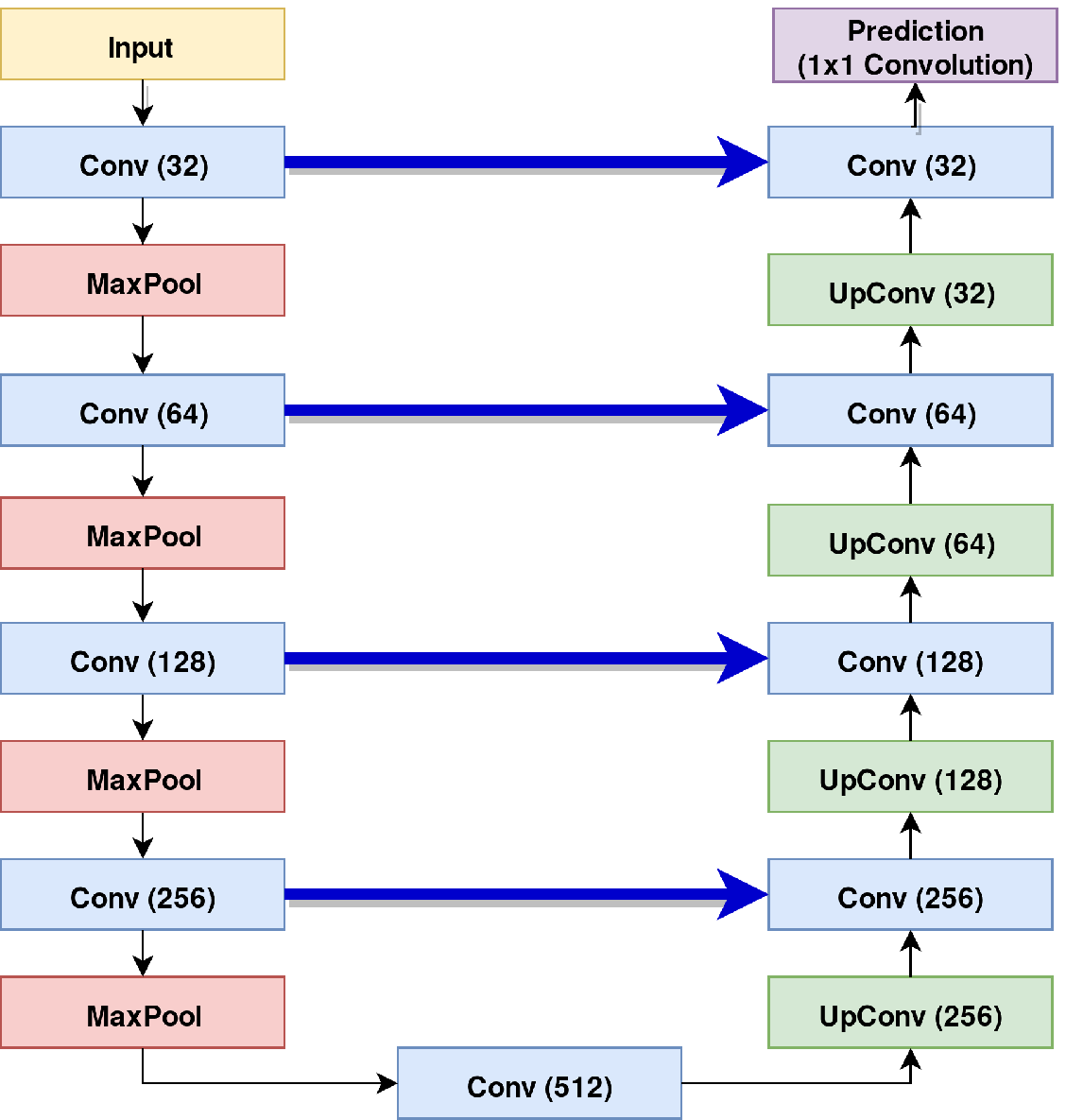}}\hfill
\subfloat[]{\includegraphics[width=0.20\textwidth,keepaspectratio]{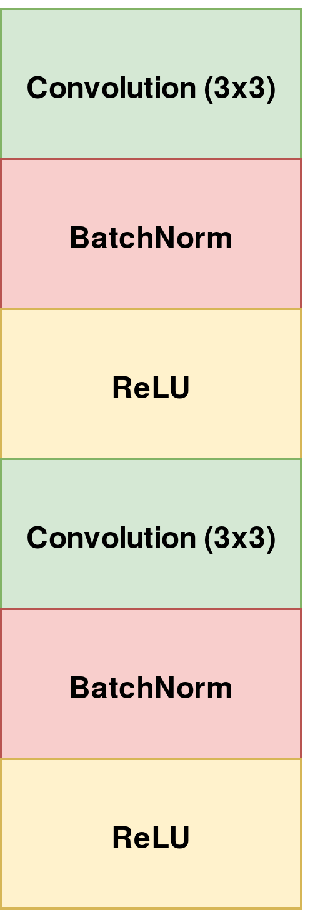}}\\

\caption{Architecture of the proposed network. (a) Proposed FCNN. (b) Composition of the \textbf{Conv} block.}
\label{network}
\end{figure}

The architecture of the network is given in Fig. (\ref{network} (a)). Each \textbf{Conv} block in the network comprises of 2 sets of convolution by 3x3 kernels, batch normalization and a non linearity (ReLU), (Fig. (\ref{network} (b))). The number of filters in each layer is given inside parenthesis in the \textbf{Conv} and \textbf{UpConv} block. The concatenation of feature maps is presented in the architecture as blue arrows.
\par The stride, kernel size \& padding  of the transposed convolution were chosen so as to produce feature maps of similar dimensions as that of the feature maps of the adjoining \textbf{Conv} block. This enables concatenation of feature maps without the need of cropping feature maps from the \textbf{Conv} block. The network makes use of convolution with 1x1 filters in the hindmost convolution block and results in generating the segmentation map.


\subsection{Training}
\par The network was trained and validated using with slices extracted from 120  and 50 HGG patients respectively. The weights and biases in each layer was initialized using the Xavier initialization \cite{xavier}. The network was trained for 30 epochs and the weights and biases were learned by minimizing the cross entropy loss function  with ADAM \cite {adam} as the optimizer. 

\par The imbalance amongst classes in the dataset  were addressed by:
\begin{enumerate}
\item Training and validating the network using slices that comprises of atleast one pixel of lesion.
\item Performing data augmentation on the extracted slices which include horizontal flipping of the data.
\item Using a weighted cross entropy as the loss function for training the network. The weight assigned to normal, necrotic, edema and enhancing were 1, 5, 2 and 3 respectively.
\end{enumerate}

\subsection{Testing}
During the testing phase, axial slices from all 4 sequences were fed to the trained network to generate the segmentation mask/ volumes.

\subsection{Post processing}

CNNs being deterministic techniques tend to mis-classify voxels as lesions at certain locations such as cerebellum, brain stem etc. were occurrence of gliomas is physiologically impossible. The false positives in predictions made by the trained network were removed by using 3-D connected component analysis. All components below a certain threshold (T=2000) were discarded while the rest were retained. 

\subsection{Survival prediction}
\begin{figure}
\subfloat[]{\includegraphics[width=0.20\textwidth,keepaspectratio]{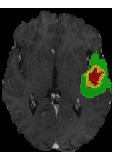}}\hfill
\subfloat[]{\includegraphics[width=0.20\textwidth,keepaspectratio]{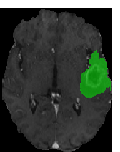}} \hfill
\subfloat[]{\includegraphics[width=0.20\textwidth,keepaspectratio]{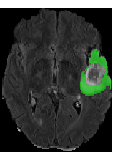}}\hfill
\subfloat[]{\includegraphics[width=0.20\textwidth,keepaspectratio]{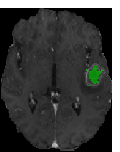}} \hfill
\subfloat[]{\includegraphics[width=0.20\textwidth,keepaspectratio]{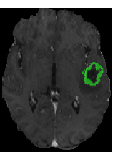}} \\
\caption{Mask generated from segmentation maps. (a) Prediction made by the segmentation network. (b) whole lesion mask. (c) edema mask. (d) necrosis mask. (e) enhancing mask. In figure (a), green, red and yellow indicate edema, necrotic core and enhancing tumor in the lesion respectively.}
\label{fig:masks}
\end{figure}

The segmentation mask generated by the network was binarized to form 4 different volumes namely whole lesion mask, edema mask, necrosis mask and enhancing mask, (Fig. \ref{fig:masks}). A total of 19 first order texture based features, Table \ref{tab:text} and 16 shape based features of the lesion, Table \ref{tab:shape}, were extracted from T1 post contrast sequences using each of the aforementioned masks. The texture and shape based features were extracted from MR volume using a python package called Pyradiomics \cite{pyradiomics}. Apart from texture and shape based features, the age of subject was used as a feature for the regressor to train and predict the prognosis.

\begin{table}[]
\centering
\caption{First order texture features extracted from T1 post contrast sequence to predict overall survival of a subject. }
\label{tab:text}
\begin{tabular}{|l|l|}
\hline
\multicolumn{2}{|c|}{First order Texture Based Feature} \\ \hline
1. Volume & 11. Range \\ \hline
2. Total Energy & 12. Mean Absolute Deviation \\ \hline
3. Entropy & 13. Robust Mean Absolute Deviation \\ \hline
4. Minimum & 14. Root Mean Squared \\ \hline
5. 10th percentile & 15. Standard Deviation \\ \hline
6. 90th percentile & 16. Skewness \\ \hline
7. Maximum & 17. Kurtosis \\ \hline
8. Mean & 18. Variance \\ \hline
9. Median & 19. Uniformity \\ \hline
10. Interquartile Range &  \\ \hline
\end{tabular}
\end{table}

\begin{table}[]
\centering
\caption{Shape based features used to predict prognosis of a subject}
\label{tab:shape}
\begin{tabular}{|l|l|}
\hline
\multicolumn{2}{|c|}{Shape Based Feature} \\ \hline
1. Volume &  9. Maximum 2D diameter (coronal) \\ \hline
2. Surface area & 10. Maximum 2D diameter (sagital) \\ \hline
3. Surface area to Volume Ratio & 11. Major Axis \\ \hline
4. Sphericity & 12. Minor Axis \\ \hline
5. Spherical Disproportion & 13. Least Axis \\ \hline
6. Compactness 1 & 14. Elongation \\ \hline
7. Maximum 3D diameter & 15. Flatness \\ \hline
8. Maximum 2D diameter (axial) & 16. Compactness 2 \\ \hline
\end{tabular}
\end{table}

\section{Results}
\begin{figure}
\subfloat[]{\includegraphics[width=0.25\textwidth,keepaspectratio]{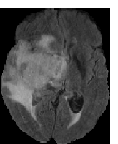}}\hfill
\subfloat[]{\includegraphics[width=0.25\textwidth,keepaspectratio]{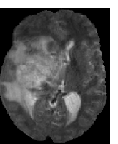}} \hfill
\subfloat[]{\includegraphics[width=0.25\textwidth,keepaspectratio]{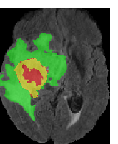}}\hfill
\subfloat[]{\includegraphics[width=0.25\textwidth,keepaspectratio]{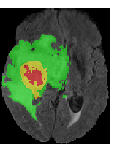}} \\
\subfloat[]{\includegraphics[width=0.25\textwidth,keepaspectratio]{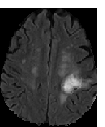}}\hfill
\subfloat[]{\includegraphics[width=0.25\textwidth,keepaspectratio]{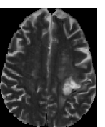}} \hfill
\subfloat[]{\includegraphics[width=0.25\textwidth,keepaspectratio]{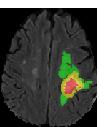}}\hfill
\subfloat[]{\includegraphics[width=0.25\textwidth,keepaspectratio]{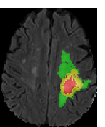}} \\
\caption{Performance of the network on local test data. (a) FLAIR. (b) T2. (c) Prediction. (d) Ground Truth. (e) FLAIR. (f) T2. (g) Prediction. (h) Ground Truth. In figures c,d,g \& h, green- Edema, yellow- Enhancing Tumor, red- Necrotic Core.}
\label{fig:results}
\end{figure}

The performance of network on the entire BraTS on the local test set (n=40) [HGG-25, LGG-15] is given in Table (\ref{local_per}). Fig.(\ref{fig:results}) shows the performance of the network on 2 different patients from the local test data.  
 
\begin{table}[]
\centering
\caption{Performance of the network on local test data (HGG=25 and LGG=15)}
\label{local_per}
\begin{tabular}{|c|c|c|c|}
\hline
       & Whole Tumor & Tumor Core & Active Tumor \\
       \hline
Mean   & 0.84        & 0.84       & 0.77         \\
\hline
Std. Dev.    & 0.19        & 0.20       & 0.19         \\
\hline
Median & 0.90        & 0.90       & 0.83        \\
\hline
\end{tabular}
\end{table}
\par
\begin{figure}
\subfloat[]{\includegraphics[width=0.25\textwidth,keepaspectratio]{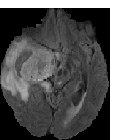}}\hfill
\subfloat[]{\includegraphics[width=0.25\textwidth,keepaspectratio]{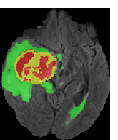}} \hfill
\subfloat[]{\includegraphics[width=0.25\textwidth,keepaspectratio]{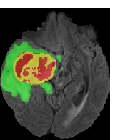}}\hfill
\subfloat[]{\includegraphics[width=0.25\textwidth,keepaspectratio]{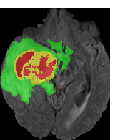}} \\
\caption{Reduction of False positive using connected components. (a) FLAIR. (b) Raw Prediction. (c) Post Processed image. (d) Ground Truth. In figures b, c, d, green- Edema, yellow- Enhancing Tumor, red- Necrotic Core.}
\label{fig:pp}
\end{figure}

 The post processing technique improves the performance of the network. On the local test data, the improvement in performance was in the order of 2.44\% for whole tumor dice score, 2.44\% for tumor core and 1.31\% for active tumor. Fig.(\ref{fig:pp}) shows an example were the proposed post processing technique aids in eliminating false positives.

\par For the task of overall survival rate prediction, it was observed that texture based \& shape based features extracted from T1 post contrast sequence performed better than extracting features from other MR sequences. We observed that using features extracted all four sequences had negative impact on performance of the regressor. 

\par The performance of the network on the BraTS 2017 validation set is given in Table (\ref{val}). It was observed that the network maintains similar whole tumor scores on the local test data and on the validation data. However, a dip in performance was observed in the tumor core \& active tumor compartments. The performance of the proposed technique for survival prediction on the validation data is given in Table (\ref{val_sur}). 

\begin{table}[H]
\centering
\caption{Performance of the network on BraTS 2017 validation data}
\label{val}
\begin{tabular}{|c|c|c|c|}
\hline
       & Whole Tumor & Tumor Core & Active Tumor \\
       \hline
Mean   & 0.83        & 0.69       & 0.72         \\
\hline
Std. Dev.    & 0.16        & 0.30       & 0.32         \\
\hline
Median & 0.90        & 0.83       & 0.85       \\
\hline
\end{tabular}
\end{table}

\begin{table}[H]
\centering
\caption{Survival prediction on BraTS 2017 validation data}
\label{val_sur}
\begin{tabular}{|c|c|c|c|c|}
\hline
       Accuracy & MSE& Median SE & Std SE & SpearmanR \\
       \hline
     0.52      & 221203.54 & 59035.10 &  505184.81 &  	0.27        \\

\hline
\end{tabular}
\end{table}

\par The trained network was tested on BraTS 2017 challenge data (n=146). The performance of the proposed algorithm for the task of segmentation of gliomas from multi modal MR images is given in Table 4. For the overall survival prediction, the proposed technique achieved an accuracy of 47\% \& and a Spearman coefficient of 0.41, (Table 5). 

\par The proposed technique which makes use of a single network, produces good segmentations on the challenge data \& its performance was found to be comparable with techniques that uses an ensemble of networks. 

\begin{table}[H]
\centering
\caption{Performance of the network on BraTS 2017 testing data (n=146)}
\label{val}
\begin{tabular}{|c|c|c|c|}
\hline
       & Whole Tumor & Tumor Core & Active Tumor \\
       \hline
Mean   & 0.79        & 0.65       & 0.63         \\
\hline
Std. Dev.    & 0.20        & 0.33       & 0.33         \\
\hline
Median & 0.87        & 0.82       & 0.77       \\
\hline
\end{tabular}
\end{table}

\begin{table}[H]
\centering
\caption{Survival prediction on BraTS 2017 testing data (n=95)}
\label{val_sur}
\begin{tabular}{|c|c|c|c|c|}
\hline
       Accuracy & MSE& Median SE & Std SE & SpearmanR \\
       \hline
     0.47      & 217755.772 & 39264.206 &  668015.465 &  	0.411        \\

\hline
\end{tabular}
\end{table}

\section{Conclusion}

In this paper, we propose a fully automatic technique for segmentation of gliomas from MR volume and predict the prognosis of the patient using first order texture and shape based features. A fully convolutional neural network was utilized for task  segmentation of gliomas into its various constituents namely edema, necrotic core and enhancing tumor. A 3-D connected component analysis was used to remove false positives in the predictions made by the network. The network produces good segmentation on the BraTS test data and achieved a whole tumor, tumor core and active tumor dice score of 0.79, 0.65 and 0.63 respectively. The segmentation produced by the network was used to generate 4 different masks namely whole tumor mask, edema mask, necrotic mask, enhancing mask.   Using each mask 19 different first order texture features and 16 shaped based features were extracted from T1 post contrast sequence to train a XGBOOST regressor to predict the prognosis of a subject. On the BraTS 2017  validation data and test data, the regressor achieved an accuracy of 52 \% and 47\%  respectively.








%
%

\end{document}